\documentclass[11pt]{article}

\usepackage[preprint]{acl}

\usepackage{times}
\usepackage{latexsym}
\usepackage[T1]{fontenc}
\usepackage[utf8]{inputenc}
\usepackage{microtype}
\usepackage{inconsolata}
\usepackage{graphicx}
\usepackage{booktabs}
\usepackage{multirow}
\usepackage{tabularx}
\usepackage{xcolor}
\usepackage{amsmath}
\usepackage{amssymb}
\usepackage{xspace}

\makeatletter
\def\BreakableUnderscore{\leavevmode\hbox{\textunderscore}\nobreak\hskip 0pt\@plus 0pt\penalty 0\relax}
\makeatother

\newcommand{\systemname}{DisasterLex\xspace}

\title{DisasterLex: An Expert Concept-to-Schema Knowledge Graph for Geospatial Reasoning in Disaster Analytics}

\renewcommand{\thefootnote}{\fnsymbol{footnote}}
\author{
  \textbf{Yiming Xiao}\thanks{Corresponding author.} \quad
  \textbf{Ankit Basu} \quad
  \textbf{Kai Yin} \\
  \textbf{Sahil Vartak} \quad
  \textbf{Christian Swords} \quad
  \textbf{Ali Mostafavi} \\
  Texas A\&M University \\
  \texttt{\{yxiao, ankitbasu, kai\_yin, svartak, c.swords.9102, mostafavi\}@tamu.edu}
}

\begin{document}
\maketitle
\renewcommand{\thefootnote}{\arabic{footnote}}
\setcounter{footnote}{0}

\begin{abstract}
Disasters are inevitable and increasingly costly, and effective response depends on querying structured tabular data: precise, information-dense records of hazard, exposure, vulnerability, and lifeline infrastructure that underpin disaster management. Current text-to-SQL methods enable natural-language access to such tables but transfer poorly to the disaster domain, where queries span heterogeneous geospatial schemas and require reasoning over causal relations. We introduce \textbf{\systemname}, a knowledge-graph-mediated framework that inserts an Expert Knowledge Graph (EKG) of curated concepts and typed causal edges between the user query and the database, bridged to schema by concept-to-table links. The orchestration runs four stages (identifying query entities, routing to the operational domain, planning over causal edges, and grounding the SQL), restricting the schema passed to the model at each step. We instantiate it on a disaster-analytics database (36 geospatial tables, 150 columns) with an EKG of 107 concepts, 117 causal edges, and 52 concept-to-schema links, evaluated on a 75-query test set. On all seven base models spanning proprietary and open-weight families, \systemname beats four state-of-the-art baselines (LightRAG, HippoRAG~2, ReFoRCE, CHESS) by 1.4$\times$ to 2.75$\times$, with absolute scores of 1.65 to 3.56 (of 5.0). Error analysis shows baseline failures cluster in routing and multi-table SQL composition, the operations our orchestration explicitly addresses. Code, data, and the EKG artifact are available at \href{https://github.com/YimingXiao98/DisasterLex}{this repository} and on Zenodo at \href{https://doi.org/10.5281/zenodo.20388029}{10.5281/zenodo.20388029}.
\end{abstract}

\section{Introduction}
\label{sec:intro}

Disasters, both natural and technological, inevitably trigger severe societal, economic, and humanitarian consequences~\citep{fan2020disaster_city, leiHarnessingLargeLanguage}. During such events, emergency analysts, incident commanders, and affected populations rely on rapid, accurate access to structured geospatial data (hazard exposure, population vulnerability, lifeline-infrastructure readiness) to coordinate response under time-critical decision constraints~\citep{fema2017nims, comfort2007interorganizational, bharosa2010challenges}. Reliable natural-language access to this data could dramatically lower the barrier between an analyst's question and the underlying tables, but the data itself resists general-purpose language tools: it spans dozens of heterogeneous tables with specialised semantics, where sentinel values mark missingness, scores invert between datasets so that higher means more resilient on one index but more vulnerable on another, and the causal relations a domain expert reasons over (e.g., low community resilience reduces emergency-response capacity, flood depth increases structural damage, power outages cascade to hospital operations) must be encoded externally for the LLM to use.

Current natural-language data access approaches fail in this setting for distinct reasons. Text-to-SQL systems~\citep{yu2018spider, li2023bird, pourreza2023dinsql} assume a small or pre-selected schema; at our scale, injecting the full 150-column schema into every Text-to-SQL prompt is technically feasible at modern context lengths but dilutes attention across semantically unrelated tables, producing hallucinated joins between, e.g., flood-risk tables and unrelated demographic tables that share only an \texttt{hex\_id} key. Standard retrieval-augmented generation (RAG)~\citep{lewis2020rag, gao2024ragsurvey} retrieves passages when what is needed is executable SQL over structured tables. Graph-augmented RAG~\citep{edge2024graphrag, peng2024graphretrieval} builds knowledge graphs from document corpora to improve passage retrieval, but does not address the schema selection problem that gates structured query answering. Multi-agent text-to-SQL pipelines such as CHESS~\citep{talaeiCHESSContextualHarnessing2024}, designed for clean BIRD-style benchmarks, similarly rely on column-name heuristics for schema linking and do not encode domain causal structure.

To bridge this gap, we introduce \textbf{\systemname}, a knowledge-graph-mediated agentic framework for structured question answering in disaster analytics. At its core is a concept-level schema-linking layer: an expert-curated graph of domain concepts and typed causal edges mapped to executable database schemas. The graph mediates between user vocabulary and column names at the table-selection step, leaving SQL generation to the LLM. A four-stage orchestration pipeline (criticality extraction, operationally-motivated routing, causal-informed planning, grounded execution) enforces the operational structure that domain experts already follow. Unlike prior graph-augmented retrieval, schema-linking, and text-to-SQL systems, \systemname uses an expert-curated causal concept graph as a domain-specific schema selection layer over structured tables, motivated by the functional structure of the target domain rather than the general-purpose retrieval setting these systems were designed for.

\paragraph{Contributions.}
\noindent \textbf{(1) Concept-level schema linking via an expert-curated typed causal graph.} We address the schema-selection failure mode of prior text-to-SQL and RAG approaches by interposing an Expert Knowledge Graph (EKG) of 107~domain concepts and 117~typed causal edges between natural-language queries and a 36-table relational schema, connected by 52 explicit concept-to-schema edges. Synonym-based concept matching plus 1-hop graph traversal reduces the prompt schema context from 150 columns to typically 10--20 per query.

\noindent \textbf{(2) An ICS-motivated four-stage orchestration pipeline whose components are individually load-bearing.} We decompose disaster-analytics queries into criticality extraction, three-cluster routing, causal-informed planning, and tool-augmented execution rather than relying on a single ReAct loop~\citep{yao2023react,fema2017nims,bigley2001ics}. Ablating routing drops Tier~M (multi-table composition) by $-$1.50 on Gemini; ablating planning drops Tier~K by $-$0.29 to $-$3.19 across models, demonstrating that orchestration choices interact non-trivially with base-model capability.

\noindent \textbf{(3) A four-tier diagnostic benchmark.} A 75-case test split decoupled from development tuning, organized into four tiers that isolate distinct failure modes of natural-language interfaces to geospatial databases (routing, EKG grounding, multi-table composition, data-availability disclosure). Per-tier scoring localizes where SOTA external systems collapse, with the largest gaps concentrated on routing and multi-table composition (Table~\ref{tab:baselines}).

\noindent \textbf{(4) Cross-model evidence with quantified seed variance.} A five-condition ablation (full pipeline plus four internal-component ablations) replayed across seven architecturally diverse base models spanning closed-source and open-source families (Gemini~3.1 Flash-Lite Preview, DeepSeek~V3.2, Qwen~3.6 Flash, Llama~3.1~8B, Qwen3~8B, Qwen3~32B, Llama~3.3~70B) with three random seeds per cell, alongside four state-of-the-art external comparators: two graph-RAG retrievers (LightRAG~\citep{guoLightRAGSimpleFast2025}, HippoRAG~2~\citep{gutiérrez2025hipporagneurobiologicallyinspiredlongterm}) and two multi-agent text-to-SQL pipelines (CHESS~\citep{talaeiCHESSContextualHarnessing2024}, ReFoRCE~\citep{deng2025reforcetexttosqlagentselfrefinement}). The full pipeline beats every general-purpose retrieval substitute and text-to-SQL competitor on every base model, but cross-model ablation deltas reveal substantial heterogeneity (e.g., DeepSeek is the most tolerant of orchestration removal, Qwen~3.6 the least), qualifying any claim that ``pipeline structure matters'' by which base model is in scope.

\section{System Architecture}
\label{sec:architecture}

\begin{figure*}[t]
  \centering
  \includegraphics[width=\textwidth]{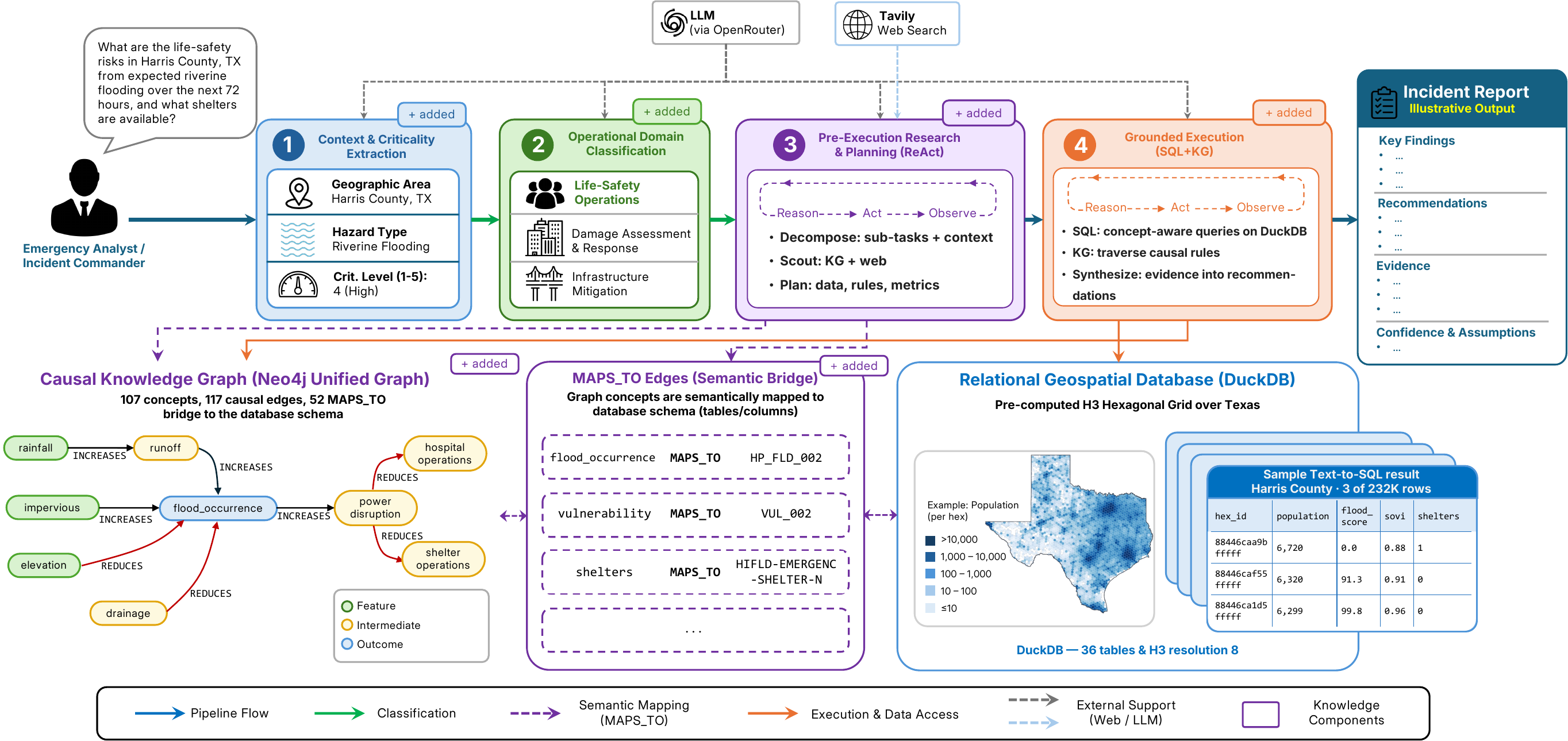}
  \caption{\textbf{\systemname architecture.} A natural-language query flows through four stages: \textbf{(1)}~context \& criticality extraction, \textbf{(2)}~operational-domain classification into a domain-specific cluster, \textbf{(3)}~a ReAct planner that scouts the EKG and the web, and \textbf{(4)}~a ReAct executor that runs concept-aware SQL on DuckDB and traverses EKG causal rules to synthesise an incident report. The expert-curated Causal Knowledge Graph is bridged to the Relational Database by \textsc{Maps\_To} edges.}
  \label{fig:architecture}
\end{figure*}

\systemname consists of three integrated components, summarized in Figure~\ref{fig:architecture}: a unified knowledge graph (\S\ref{sec:kg}), a concept-aware schema retrieval mechanism (\S\ref{sec:retrieval}), and a four-stage orchestration pipeline (\S\ref{sec:pipeline}).

\subsection{Unified Knowledge Graph}
\label{sec:kg}

The unified graph is hosted in Neo4j~\citep{webber2012neo4j} and combines a faithful representation of the relational schema with curated domain causal knowledge.

\paragraph{Disaster Data Catalog Graph (DDCG).}
Auto-introspected from a relational database: each table is a \texttt{DataTable} node with child \texttt{DataColumn} nodes, and \texttt{JoinRule} nodes encode valid join patterns over the primary-key column(s) of each table. In our case study (\S\ref{sec:evaluation}), the DDCG is auto-introspected from a DuckDB~\citep{raasveldt2019duckdb} build of 36 geospatial tables (150 columns) on an H3 (Hierarchical Hexagonal) grid~\citep{brodsky2018h3, sahr2003geodesic} at resolution~8 joined on a shared \texttt{hex\_id} key.

\paragraph{Expert Knowledge Graph (EKG).}
A directed graph of Concept nodes connected by typed causal edges (types: \textsc{Increases}, \textsc{Reduces}, \textsc{Indicates}, \textsc{Requires}, \textsc{Scales}), each carrying a confidence weight in $[0,1]$. Each concept carries a synonym list supporting fuzzy matching from natural language. In our case study, the EKG has 107 concept nodes and 117 edges spanning eight conceptual categories (features, intermediate processes, outcomes, domain anchors, interventions, exposures, response assets, infrastructure) with $\sim$85 aliases.

\paragraph{Concept-to-schema bridge.}
A set of 52 edges connects EKG concept nodes to DDCG data tables that are loaded in the current DuckDB build. For example, \texttt{flood\_occurrence} maps to \texttt{HP\_FLD\_002}, the National Risk Index riverine flood risk table~\citep{fema2023nri}, and \texttt{hospitals} maps to \texttt{EX\_LIFE\_004}, the hospital facility-count table.

\paragraph{EKG curation process.}
The EKG is developed iteratively against domain literature; each causal edge requires either an explicit cited source or a clearly entailed mechanism (e.g., \texttt{impervious}~\textsc{Increases}~\texttt{runoff}, grounded in hydrologic-process literature~\citep{Schueler1995TheIO}). Synonym lists are bootstrapped from LLM-suggested aliases and manually filtered against column documentation; validation uses trace-level spot checks against the benchmark. The case-study source list (drawn from NIMS~\citep{fema2017nims}, NRI methodology~\citep{fema2023nri}, and peer-reviewed flood, hurricane, social-vulnerability, and community-resilience literature) is released with the artifact.
\subsection{Concept-Aware Schema Retrieval}
\label{sec:retrieval}

Schema retrieval translates a natural-language query into the subset of database schemas needed for SQL generation. The procedure has three steps: (i) synonym-based concept matching against all EKG concept nodes using token-boundary regular expressions, with longest-synonym-first preference; (ii) Cypher traversal of concept-to-schema edges from activated concepts to \texttt{DataTable} nodes, retrieving columns, join rules, and data-quality warnings (a crosswalk table is always included to support cross-table joins); (iii) injection of the resulting compact schema block into the SQL generation prompt in place of the full schema. In our case study, this typically yields 10--20 columns against 150 under naive injection. For example, given the case-study query ``How many hospitals in a given county are in flood-exposed zones?'', concept matching activates \texttt{flood\_occurrence} and \texttt{hospitals}; traversal returns \texttt{HP\_FLD\_002}, \texttt{EX\_LIFE\_004}, and the crosswalk (12~columns total).

\subsection{Pipeline Orchestration}
\label{sec:pipeline}

The orchestrator is a four-stage pipeline implemented as a directed graph in LangGraph (one node per stage). 

(1)~\textbf{Context extraction} parses query context (area of interest, hazard or topic, 1--5 criticality level; criticality $\geq 3$ gates high-criticality recommendations) and runs a data-availability check that surfaces missing or blocked tables. The check is rule-based: curated sentinel-value patterns (e.g.\ \texttt{sovi} = $-999$ for missing social-vulnerability data in our case study) and a forbidden-table list are matched against the concept-to-table activation set, and any hit produces an explicit disclosure passed to downstream stages.

(2)~\textbf{Cluster routing} classifies the query into one of $k$ operational clusters specific to the domain, each with a specialized prompt template. In our case study, the three clusters (life-safety operations, damage assessment and response, infrastructure mitigation) follow the Incident Command System~\citep{fema2017nims, bigley2001ics}; the explicit ICS mapping is in Appendix~\ref{app:ics_mapping}.

(3)~\textbf{Causal-informed planning} is a ReAct~\citep{yao2023react} agent that retrieves causal edges from the EKG and produces a structured analysis plan.

(4)~\textbf{Tool-augmented execution} is a second ReAct agent with a Text-to-SQL tool (concept-aware schema retrieval, LLM-based SQL generation, syntax validation that rejects \texttt{DELETE}/\texttt{DROP} and references to blocked tables, DuckDB execution, 3-attempt retry on validation or execution errors) and a knowledge-graph tool (1-hop edge retrieval and multi-hop Cypher traversal up to 3 hops).

\section{Evaluation Design}
\label{sec:evaluation}

\paragraph{Case study and dataset.}
We use a Texas-wide disaster-analytics database as the case study to validate the framework. The DDCG is auto-introspected from a DuckDB build of 36 geospatial tables (150 columns) on the H3~\citep{brodsky2018h3} hexagonal grid at resolution~8 (827{,}648 cells per table, $\sim$0.74\,km$^2$ per cell), with a shared \texttt{hex\_id} primary key plus a county/state/ZIP crosswalk. Tables span hazard profiles (flood, hurricane, tornado, wildfire), exposure (population, critical infrastructure), social vulnerability and community resilience indices, and Homeland Infrastructure Foundation-Level Data (HIFLD) inventories. The case-study EKG and concept-to-schema mappings are described in Appendix~\ref{app:ekg}.

\subsection{Benchmark and Splits}
\label{sec:benchmark}

We evaluate on a 75-case test split organized as a four-tier taxonomy of distinct failure modes (Table~\ref{tab:tiers}). We authored the test split in two phases (county-scoped cases against the catalog of available tables, then a statewide pass adding 15 multi-county cases) and reviewed against the available DDCG schema. The test split divides into 60 county-scoped and 15 statewide queries; per-tier example cases are listed in Appendix~\ref{app:tier_examples}.

\begin{table*}[!t]
  \centering
  \small
  \renewcommand{\arraystretch}{1.15}
  \begin{tabularx}{\textwidth}{@{}l c >{\raggedright\arraybackslash}X@{}}
    \toprule
    \textbf{Tier} & \textbf{$n$} & \textbf{What it measures} \\
    \midrule
    R (routing)      & 17 & Correct ICS cluster and query-type selection (state-level check). \\
    K (EKG grounding)& 19 & Whether the expected EKG causal tokens (e.g.\ \texttt{impervious}~|~\textsc{Increases}~|~\texttt{runoff}) appear in the answer. Gold labels are drawn from the EKG, so this tier measures graph grounding rather than independent causal correctness. Web retrieval disabled. \\
    M (multi-table)  & 26 & SQL joins across 2--4 tables, with numeric tolerance bands ($\pm$100 hex counts, $\pm$10\% averages). \\
    D (disclosure)   & 13 & Transparent disclosure when data is missing, blocked, or contains sentinel values. \\
    \bottomrule
  \end{tabularx}
  \caption{Four-tier evaluation taxonomy. Each tier targets a distinct failure mode and uses tier-specific gold-label sources.}
  \label{tab:tiers}
\end{table*}

\paragraph{Scoring rubric.}
Each case produces a 1--5 judge score. Up to 4 of the 5 points come from deterministic rule-based checks against frozen gold facts; the remaining 1 point comes from an LLM-judged reasoning-quality check on the 5 statewide Tier~K cases (Gemini~2.5~Flash; prompt in Appendix~\ref{app:tierk_judge}). Per tier: Tier~R is scored by routing-state verification against the gold cluster and query-type labels; Tier~K combines entity-set matching for expected causal tokens with the reasoning judge on the statewide subset; Tier~M uses numeric value matching with tolerance bands ($\pm$100 hex counts, $\pm$10\% on averages); Tier~D uses boolean disclosure checks against gold missing-data flags.

\paragraph{Faithfulness controls.}
We bound judge-model bias and test-distribution overfitting with three controls: (i)~claim extraction and the reasoning judge both use Gemini~2.5~Flash, which is never a pipeline under evaluation, mitigating self-evaluation bias~\citep{zheng2023judging}; (ii)~4 of 5 points per case are rule-based, leaving at most 1 point of judge-model exposure; (iii)~the 75-case test split was frozen before any pipeline execution, and a separate equally sized development split with matched tier and scope distribution was used for prompt iteration (its scores are not reported). Claim extraction is validated against human annotation (95\% precision; protocol in Appendix~\ref{app:annotation}).

\subsection{Conditions}
\label{sec:conditions}

We evaluate the full \systemname pipeline against four internal ablations and four external baselines (Table~\ref{tab:conditions}). All conditions are run end-to-end on the 75-case test split across all seven base models with three random seeds per cell, with no domain-specific adaptation.

\paragraph{Ablation study.}
We isolate individual orchestration components to test their contribution to the overall pipeline.
\textbf{No Routing} replaces context extraction and cluster classification with hardcoded defaults (life-safety/shelter-placement template, criticality~3), testing whether operationally-motivated routing contributes beyond a fixed template.
\textbf{No Plan} skips the pre-execution research step, testing whether causal-informed planning adds value over a single execute-stage ReAct loop.
\textbf{ReAct} is a standard ReAct agent~\citep{yao2023react} with the same Text-to-SQL and KG tools but no routing, templates, or planning, testing whether the four-stage structure adds value over a vanilla tool-augmented agent.
\textbf{Text-RAG} replaces the EKG with chunk retrieval over the same source corpus, testing whether the EKG's typed causal structure adds value over flat retrieval of the same underlying text.

\paragraph{External baselines.}
We test state-of-the-art alternatives to the curated EKG and fall into two groups.
\textbf{LightRAG}~\citep{guoLightRAGSimpleFast2025} (auto-extracted entity-relation graph with hybrid retrieval) and \textbf{HippoRAG~2}~\citep{gutiérrez2025hipporagneurobiologicallyinspiredlongterm} (memory-style retriever using personalized PageRank over an open-relation graph) test whether general-purpose graph structure suffices once the curated EKG is removed.
\textbf{CHESS}~\citep{talaeiCHESSContextualHarnessing2024} and \textbf{ReFoRCE}~\citep{deng2025reforcetexttosqlagentselfrefinement} are multi-agent text-to-SQL pipelines, SOTA on BIRD and Spider~2.0 respectively; they test whether strong text-to-SQL agents can subsume the concept-to-schema orchestration.
All four are evaluated under their authors' recommended configurations; adapting any of them with hand-curated concept-to-schema hints would amount to porting our contribution into their architectures.

\section{Results}
\label{sec:results}

\begin{table*}[!t]
  \centering
  \footnotesize
  \renewcommand{\arraystretch}{0.95}
  \setlength{\tabcolsep}{4pt}
  \begin{tabular}{l l c c c c c}
    \toprule
    \textbf{Model} & \textbf{System} & \textbf{Overall} & \textbf{R}\textsuperscript{$\dagger$} & \textbf{K}\textsuperscript{$\dagger$} & \textbf{M}\textsuperscript{$\dagger$} & \textbf{D}\textsuperscript{$\dagger$} \\
    \midrule
    \multirow{5}{*}{\shortstack[l]{Gemini 3.1 Flash\\ Lite Preview}}
        & \textbf{\systemname (ours)} & \textbf{3.542\,$\pm$\,0.036} & \textbf{3.676\,$\pm$\,0.204} & \textbf{3.607\,$\pm$\,0.135} & \textbf{2.952\,$\pm$\,0.107} & \textbf{4.449\,$\pm$\,0.173} \\
        & LightRAG & 1.286\,$\pm$\,0.050 & 0.676\,$\pm$\,0.067 & 1.965\,$\pm$\,0.098 & 0.388\,$\pm$\,0.055 & 2.885\,$\pm$\,0.102 \\
        & ReFoRCE        & 0.620\,$\pm$\,0.054 & 0.157\,$\pm$\,0.068 & 0.474\,$\pm$\,0.139 & 0.455\,$\pm$\,0.100 & 1.769\,$\pm$\,0.133 \\
        & HippoRAG 2     & 0.502\,$\pm$\,0.004 & 0.000\,$\pm$\,0.000 & 0.377\,$\pm$\,0.015 & 0.058\,$\pm$\,0.000 & 2.231\,$\pm$\,0.000 \\
        & CHESS          & 0.417\,$\pm$\,0.195 & 0.275\,$\pm$\,0.136 & 0.456\,$\pm$\,0.237 & 0.279\,$\pm$\,0.083 & 0.821\,$\pm$\,0.451 \\
    \midrule
    \multirow{5}{*}{DeepSeek~V3.2}
        & \textbf{\systemname (ours)} & \textbf{3.033\,$\pm$\,0.083} & \textbf{3.044\,$\pm$\,0.167} & \textbf{3.235\,$\pm$\,0.051} & \textbf{2.292\,$\pm$\,0.131} & \textbf{4.256\,$\pm$\,0.118} \\
        & LightRAG & 1.123\,$\pm$\,0.058 & 0.588\,$\pm$\,0.135 & 1.534\,$\pm$\,0.172 & 0.221\,$\pm$\,0.017 & 3.026\,$\pm$\,0.291 \\
        & HippoRAG 2     & 0.651\,$\pm$\,0.027 & 0.088\,$\pm$\,0.000 & 0.710\,$\pm$\,0.066 & 0.090\,$\pm$\,0.022 & 2.423\,$\pm$\,0.168 \\
        & ReFoRCE        & 0.503\,$\pm$\,0.088 & 0.069\,$\pm$\,0.061 & 0.456\,$\pm$\,0.106 & 0.349\,$\pm$\,0.098 & 1.444\,$\pm$\,0.206 \\
        & CHESS          & 0.434\,$\pm$\,0.017 & 0.235\,$\pm$\,0.000 & 0.184\,$\pm$\,0.053 & 0.260\,$\pm$\,0.033 & 1.410\,$\pm$\,0.089 \\
    \midrule
    \multirow{5}{*}{\shortstack[l]{Qwen 3.6\\ Flash}}
        & \textbf{\systemname (ours)} & \textbf{3.555\,$\pm$\,0.114} & \textbf{3.554\,$\pm$\,0.211} & \textbf{3.679\,$\pm$\,0.141} & \textbf{2.962\,$\pm$\,0.051} & \textbf{4.564\,$\pm$\,0.178} \\
        & LightRAG & 1.407\,$\pm$\,0.117 & 0.941\,$\pm$\,0.111 & 1.801\,$\pm$\,0.096 & 0.397\,$\pm$\,0.047 & 3.462\,$\pm$\,0.428 \\
        & HippoRAG 2     & 0.591\,$\pm$\,0.040 & 0.147\,$\pm$\,0.025 & 0.412\,$\pm$\,0.076 & 0.038\,$\pm$\,0.067 & 2.538\,$\pm$\,0.000 \\
        & CHESS          & 0.462\,$\pm$\,0.014 & 0.588\,$\pm$\,0.000 & 0.325\,$\pm$\,0.030 & 0.442\,$\pm$\,0.019 & 0.538\,$\pm$\,0.000 \\
        & ReFoRCE        & 0.354\,$\pm$\,0.027 & 0.000\,$\pm$\,0.000 & 0.409\,$\pm$\,0.102 & 0.080\,$\pm$\,0.039 & 1.282\,$\pm$\,0.089 \\
    \midrule
    \multirow{5}{*}{Llama 3.1 8B}
        & \textbf{\systemname (ours)} & \textbf{1.753\,$\pm$\,0.073} & \textbf{1.864\,$\pm$\,0.149} & 1.405\,$\pm$\,0.265 & \textbf{1.394\,$\pm$\,0.076} & \textbf{3.030\,$\pm$\,0.301} \\
        & LightRAG & 1.268\,$\pm$\,0.100 & 0.824\,$\pm$\,0.025 & \textbf{1.945\,$\pm$\,0.244} & 0.660\,$\pm$\,0.024 & 2.077\,$\pm$\,0.335 \\
        & HippoRAG 2     & 1.182\,$\pm$\,0.055 & 0.559\,$\pm$\,0.092 & 1.759\,$\pm$\,0.128 & 0.260\,$\pm$\,0.035 & 3.000\,$\pm$\,0.133 \\
        & CHESS          & 0.238\,$\pm$\,0.027 & 0.069\,$\pm$\,0.061 & 0.360\,$\pm$\,0.055 & 0.032\,$\pm$\,0.022 & 0.692\,$\pm$\,0.077 \\
        & ReFoRCE        & 0.212\,$\pm$\,0.099 & 0.000\,$\pm$\,0.000 & 0.088\,$\pm$\,0.076 & 0.022\,$\pm$\,0.024 & 1.051\,$\pm$\,0.437 \\
    \midrule
    \multirow{5}{*}{Qwen3 8B}
        & \textbf{\systemname (ours)} & \textbf{2.566\,$\pm$\,0.029} & \textbf{2.598\,$\pm$\,0.189} & \textbf{2.492\,$\pm$\,0.294} & \textbf{1.920\,$\pm$\,0.063} & \textbf{3.923\,$\pm$\,0.333} \\
        & LightRAG & 1.750\,$\pm$\,0.023 & 1.044\,$\pm$\,0.092 & 2.484\,$\pm$\,0.148 & 0.679\,$\pm$\,0.047 & 3.744\,$\pm$\,0.178 \\
        & HippoRAG 2     & 0.907\,$\pm$\,0.054 & 0.059\,$\pm$\,0.051 & 1.389\,$\pm$\,0.194 & 0.269\,$\pm$\,0.035 & 2.590\,$\pm$\,0.118 \\
        & ReFoRCE\textsuperscript{$\ddagger$} & 0.330\,$\pm$\,0.024 & 0.294\,$\pm$\,0.083 & 0.289\,$\pm$\,0.112 & 0.317\,$\pm$\,0.041 & 0.462\,$\pm$\,0.326 \\
        & CHESS          & 0.069\,$\pm$\,0.039 & 0.000\,$\pm$\,0.000 & 0.070\,$\pm$\,0.066 & 0.096\,$\pm$\,0.084 & 0.103\,$\pm$\,0.118 \\
    \midrule
    \multirow{5}{*}{Qwen3 32B}
        & \textbf{\systemname (ours)} & \textbf{2.565\,$\pm$\,0.135} & \textbf{2.657\,$\pm$\,0.074} & \textbf{2.466\,$\pm$\,0.372} & \textbf{1.974\,$\pm$\,0.068} & \textbf{3.769\,$\pm$\,0.231} \\
        & LightRAG & 1.388\,$\pm$\,0.046 & 0.632\,$\pm$\,0.067 & 2.246\,$\pm$\,0.128 & 0.538\,$\pm$\,0.085 & 2.821\,$\pm$\,0.118 \\
        & HippoRAG 2     & 1.040\,$\pm$\,0.037 & 0.176\,$\pm$\,0.000 & 1.439\,$\pm$\,0.110 & 0.340\,$\pm$\,0.022 & 2.987\,$\pm$\,0.059 \\
        & ReFoRCE        & 0.516\,$\pm$\,0.066 & 0.314\,$\pm$\,0.136 & 0.237\,$\pm$\,0.091 & 0.545\,$\pm$\,0.089 & 1.128\,$\pm$\,0.311 \\
        & CHESS          & 0.242\,$\pm$\,0.213 & 0.078\,$\pm$\,0.068 & 0.123\,$\pm$\,0.106 & 0.147\,$\pm$\,0.128 & 0.821\,$\pm$\,0.747 \\
    \midrule
    \multirow{5}{*}{Llama 3.3 70B}
        & \textbf{\systemname (ours)} & \textbf{1.650\,$\pm$\,0.133} & \textbf{1.662\,$\pm$\,0.092} & 1.151\,$\pm$\,0.189 & \textbf{1.346\,$\pm$\,0.173} & \textbf{2.974\,$\pm$\,0.160} \\
        & HippoRAG 2     & 1.200\,$\pm$\,0.060 & 0.338\,$\pm$\,0.067 & \textbf{2.026\,$\pm$\,0.040} & 0.324\,$\pm$\,0.034 & 2.872\,$\pm$\,0.311 \\
        & LightRAG & 1.076\,$\pm$\,0.152 & 0.809\,$\pm$\,0.135 & 1.706\,$\pm$\,0.387 & 0.538\,$\pm$\,0.108 & 1.577\,$\pm$\,0.390 \\
        & ReFoRCE        & 0.499\,$\pm$\,0.184 & 0.196\,$\pm$\,0.068 & 0.351\,$\pm$\,0.219 & 0.285\,$\pm$\,0.147 & 1.538\,$\pm$\,0.407 \\
        & CHESS          & 0.158\,$\pm$\,0.148 & 0.039\,$\pm$\,0.068 & 0.061\,$\pm$\,0.106 & 0.103\,$\pm$\,0.111 & 0.564\,$\pm$\,0.395 \\
    \bottomrule
  \end{tabular}
  \\[3pt]
  \begin{minipage}{\linewidth}\raggedright
  \textsuperscript{$\dagger$} R: routing (cluster + query type); K: EKG grounding; M: multi-table SQL composition; D: data-availability disclosure (\S\ref{sec:benchmark}). External baselines run end-to-end in their authors' designed mode (no routing / EKG / DB access), scored on the same harness. \textsuperscript{$\ddagger$} ReFoRCE / Qwen3~8B: $n=2$ seeds (seed~2 exceeded OpenRouter's 1\,M-token context window).
  \end{minipage}
  \caption{\textbf{\systemname vs.\ four external baselines on the 75-case test split.} Mean $\pm$ std over 3 seeds (fractional LLM judge, 0--5); within each model group, baselines are sorted by overall score. Claim extraction uses Gemini~2.5~Flash to prevent self-evaluation bias. Seven base models spanning closed-source and open-source families (8B--70B parameters).}
  \label{tab:baselines}
\end{table*}

\begin{table*}[!t]
  \centering
  \footnotesize
  \renewcommand{\arraystretch}{0.95}
  \setlength{\tabcolsep}{4pt}
  \begin{tabular}{l l c c c c c}
    \toprule
    \textbf{Model} & \textbf{Condition} & \textbf{Overall} & \textbf{R}\textsuperscript{$\dagger$} & \textbf{K}\textsuperscript{$\dagger$} & \textbf{M}\textsuperscript{$\dagger$} & \textbf{D}\textsuperscript{$\dagger$} \\
    \midrule
    \multirow{5}{*}{\shortstack[l]{Gemini 3.1 Flash\\ Lite Preview}}
        & \textbf{\systemname (ours)}  & \textbf{3.542\,$\pm$\,0.036} & \textbf{3.676\,$\pm$\,0.204} & \textbf{3.607\,$\pm$\,0.135} & \textbf{2.952\,$\pm$\,0.107} & \textbf{4.449\,$\pm$\,0.173} \\
        & No Plan      & 2.955\,$\pm$\,0.425 & 3.039\,$\pm$\,0.514 & 3.132\,$\pm$\,0.296 & 2.428\,$\pm$\,0.354 & 4.448\,$\pm$\,0.303 \\
        & No Routing   & 2.489\,$\pm$\,0.021 & 2.402\,$\pm$\,0.180 & 2.824\,$\pm$\,0.041 & 1.455\,$\pm$\,0.048 & 4.179\,$\pm$\,0.222 \\
        & ReAct        & 2.404\,$\pm$\,0.046 & 2.054\,$\pm$\,0.122 & 2.772\,$\pm$\,0.026 & 1.811\,$\pm$\,0.053 & 3.513\,$\pm$\,0.097 \\
        & Text-RAG     & 2.396\,$\pm$\,0.076 & 2.221\,$\pm$\,0.053 & 2.690\,$\pm$\,0.149 & 1.795\,$\pm$\,0.015 & 3.397\,$\pm$\,0.349 \\
    \midrule
    \multirow{5}{*}{DeepSeek~V3.2}
        & \textbf{\systemname (ours)}  & \textbf{3.033\,$\pm$\,0.083} & \textbf{3.044\,$\pm$\,0.167} & \textbf{3.235\,$\pm$\,0.051} & \textbf{2.292\,$\pm$\,0.131} & 4.256\,$\pm$\,0.118 \\
        & No Plan      & 2.868\,$\pm$\,0.011 & 2.765\,$\pm$\,0.102 & 2.947\,$\pm$\,0.099 & 2.272\,$\pm$\,0.056 & 4.077\,$\pm$\,0.154 \\
        & No Routing   & 2.231\,$\pm$\,0.057 & 1.936\,$\pm$\,0.066 & 2.372\,$\pm$\,0.380 & 1.240\,$\pm$\,0.183 & \textbf{4.393\,$\pm$\,0.393} \\
        & ReAct        & 2.660\,$\pm$\,0.044 & 2.424\,$\pm$\,0.058 & 2.955\,$\pm$\,0.071 & 2.003\,$\pm$\,0.064 & 3.910\,$\pm$\,0.308 \\
        & Text-RAG     & 2.496\,$\pm$\,0.067 & 2.364\,$\pm$\,0.106 & 2.816\,$\pm$\,0.107 & 1.790\,$\pm$\,0.094 & 3.944\,$\pm$\,0.255 \\
    \midrule
    \multirow{5}{*}{\shortstack[l]{Qwen 3.6\\ Flash}}
        & \textbf{\systemname (ours)}  & \textbf{3.555\,$\pm$\,0.114} & \textbf{3.554\,$\pm$\,0.211} & \textbf{3.679\,$\pm$\,0.141} & \textbf{2.962\,$\pm$\,0.051} & \textbf{4.564\,$\pm$\,0.178} \\
        & ReAct        & 2.448\,$\pm$\,0.124 & 2.255\,$\pm$\,0.177 & 2.524\,$\pm$\,0.100 & 1.849\,$\pm$\,0.209 & 4.281\,$\pm$\,0.267 \\
        & Text-RAG     & 2.316\,$\pm$\,0.104 & 2.241\,$\pm$\,0.240 & 2.715\,$\pm$\,0.113 & 1.683\,$\pm$\,0.043 & 3.496\,$\pm$\,0.277 \\
        & No Plan      & 1.440\,$\pm$\,0.045 & 1.397\,$\pm$\,0.082 & 0.487\,$\pm$\,0.173 & 1.115\,$\pm$\,0.084 & 3.538\,$\pm$\,0.154 \\
        & No Routing   & 0.804\,$\pm$\,0.246 & 0.824\,$\pm$\,0.025 & 0.421\,$\pm$\,0.194 & 0.290\,$\pm$\,0.061 & 2.159\,$\pm$\,1.439 \\
    \midrule
    \multirow{5}{*}{Llama 3.1 8B}
        & \textbf{\systemname (ours)}  & 1.753\,$\pm$\,0.073 & 1.864\,$\pm$\,0.149 & 1.405\,$\pm$\,0.265 & 1.394\,$\pm$\,0.076 & \textbf{3.030\,$\pm$\,0.301} \\
        & No Routing   & 0.980\,$\pm$\,0.099 & 0.912\,$\pm$\,0.178 & 0.588\,$\pm$\,0.151 & 0.490\,$\pm$\,0.020 & 2.611\,$\pm$\,0.530 \\
        & No Plan      & 1.654\,$\pm$\,0.105 & 1.842\,$\pm$\,0.051 & 1.210\,$\pm$\,0.319 & 1.420\,$\pm$\,0.200 & 2.679\,$\pm$\,0.680 \\
        & ReAct        & 1.787\,$\pm$\,0.064 & 1.809\,$\pm$\,0.156 & \textbf{2.161\,$\pm$\,0.106} & \textbf{1.499\,$\pm$\,0.056} & 2.225\,$\pm$\,0.461 \\
        & Text-RAG     & \textbf{1.801\,$\pm$\,0.097} & \textbf{1.900\,$\pm$\,0.109} & 1.828\,$\pm$\,0.084 & 1.352\,$\pm$\,0.075 & 2.981\,$\pm$\,0.694 \\
    \midrule
    \multirow{5}{*}{Qwen3 8B}
        & \textbf{\systemname (ours)}  & \textbf{2.566\,$\pm$\,0.029} & \textbf{2.598\,$\pm$\,0.189} & 2.492\,$\pm$\,0.294 & 1.920\,$\pm$\,0.063 & 3.923\,$\pm$\,0.333 \\
        & No Routing   & 1.624\,$\pm$\,0.044 & 1.397\,$\pm$\,0.092 & 1.689\,$\pm$\,0.187 & 0.603\,$\pm$\,0.089 & 3.868\,$\pm$\,0.396 \\
        & No Plan      & 2.336\,$\pm$\,0.144 & 2.260\,$\pm$\,0.114 & 2.243\,$\pm$\,0.270 & 1.788\,$\pm$\,0.178 & 3.667\,$\pm$\,0.347 \\
        & ReAct        & 2.226\,$\pm$\,0.077 & 2.218\,$\pm$\,0.154 & \textbf{2.633\,$\pm$\,0.231} & \textbf{1.939\,$\pm$\,0.063} & \textbf{4.139\,$\pm$\,0.438} \\
        & Text-RAG     & 1.854\,$\pm$\,0.029 & 2.501\,$\pm$\,0.093 & 1.320\,$\pm$\,0.128 & 1.543\,$\pm$\,0.082 & 3.908\,$\pm$\,0.119 \\
    \midrule
    \multirow{5}{*}{Qwen3 32B}
        & \textbf{\systemname (ours)}  & \textbf{2.565\,$\pm$\,0.135} & \textbf{2.657\,$\pm$\,0.074} & 2.466\,$\pm$\,0.372 & \textbf{1.974\,$\pm$\,0.068} & 3.769\,$\pm$\,0.231 \\
        & No Routing   & 1.737\,$\pm$\,0.067 & 1.721\,$\pm$\,0.025 & 1.671\,$\pm$\,0.388 & 0.721\,$\pm$\,0.038 & \textbf{3.889\,$\pm$\,0.175} \\
        & No Plan      & 2.163\,$\pm$\,0.046 & 1.995\,$\pm$\,0.081 & 2.017\,$\pm$\,0.140 & 1.603\,$\pm$\,0.119 & 3.718\,$\pm$\,0.194 \\
        & ReAct        & 2.152\,$\pm$\,0.207 & 2.280\,$\pm$\,0.207 & \textbf{2.857\,$\pm$\,0.087} & 1.921\,$\pm$\,0.135 & 3.648\,$\pm$\,0.306 \\
        & Text-RAG     & 2.106\,$\pm$\,0.098 & 2.304\,$\pm$\,0.111 & 2.792\,$\pm$\,0.115 & 1.717\,$\pm$\,0.140 & 3.448\,$\pm$\,0.390 \\
    \midrule
    \multirow{5}{*}{Llama 3.3 70B}
        & \textbf{\systemname (ours)}  & 1.650\,$\pm$\,0.133 & \textbf{1.662\,$\pm$\,0.092} & 1.151\,$\pm$\,0.189 & 1.346\,$\pm$\,0.173 & 2.974\,$\pm$\,0.160 \\
        & No Routing   & 1.150\,$\pm$\,0.116 & 0.647\,$\pm$\,0.051 & 1.180\,$\pm$\,0.339 & 0.532\,$\pm$\,0.031 & 3.000\,$\pm$\,0.277 \\
        & No Plan      & 1.659\,$\pm$\,0.075 & 1.441\,$\pm$\,0.135 & 1.298\,$\pm$\,0.053 & 1.369\,$\pm$\,0.135 & \textbf{3.051\,$\pm$\,0.118} \\
        & ReAct        & \textbf{1.667\,$\pm$\,0.140} & 1.642\,$\pm$\,0.132 & 1.833\,$\pm$\,0.249 & 1.253\,$\pm$\,0.195 & 2.286\,$\pm$\,0.091 \\
        & Text-RAG     & 1.666\,$\pm$\,0.054 & 1.299\,$\pm$\,0.073 & \textbf{1.930\,$\pm$\,0.033} & \textbf{1.474\,$\pm$\,0.034} & 2.145\,$\pm$\,0.327 \\
    \bottomrule
  \end{tabular}
  \\[3pt]
  \begin{minipage}{\linewidth}\raggedright
  \textsuperscript{$\dagger$} Tiers as in Table~\ref{tab:baselines}.
  \end{minipage}
  \caption{\textbf{Internal-component ablation study.} Full \systemname pipeline vs.\ four ablations: \textbf{No Plan} (skip planning), \textbf{No Routing} (skip routing), \textbf{ReAct} (vanilla ReAct with the same tools), \textbf{Text-RAG} (chunk retrieval in place of the EKG). Seven base models (8B--70B). Llama~3.1~8B used \texttt{response\_format=json\_object} on Stages~1--2; Qwen3~32B ReAct ran with a 600\,s per-case budget.}
  \label{tab:ablation}
\end{table*}

\paragraph{Main results.}
The full \systemname pipeline scores \textbf{1.65--3.56} overall across seven base models on the 75-case test split, with three random seeds per cell (Gemini~3.1 Flash-Lite Preview, DeepSeek~V3.2, Qwen~3.6 Flash, Llama~3.1~8B, Qwen3~8B, Qwen3~32B, Llama~3.3~70B; Table~\ref{tab:baselines}). The per-tier shape (Tier~D strongest, Tier~M weakest) is stable across all seven base models. Sensitivity to internal-component ablation, however, varies substantially by base model (Table~\ref{tab:ablation}); we unpack the patterns below.

\begin{figure*}[!htbp]
  \centering
  \includegraphics[width=\textwidth]{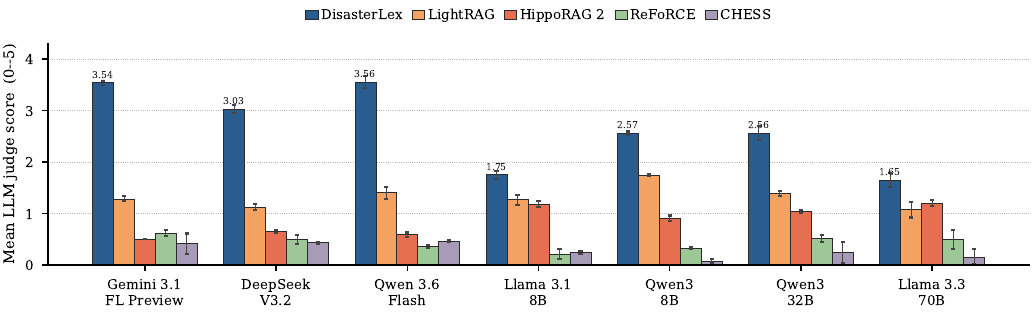}
  \caption{Overall mean LLM judge score on the 75-case test split for \systemname vs.\ four external baselines (LightRAG, HippoRAG~2, ReFoRCE, CHESS) across all seven base models. DisasterLex Full bars are value-labeled. Bars are means over three random seeds; error bars are $\pm$ one standard deviation.}
  \label{fig:cross_model_baselines}
\end{figure*}

\paragraph{External baselines fall well below the full pipeline on every model.}
The four state-of-the-art external systems (LightRAG, HippoRAG~2, ReFoRCE, CHESS) score \textbf{1.4--2.75$\times$ below} \systemname on every base model (Table~\ref{tab:baselines}, Figure~\ref{fig:cross_model_baselines}). LightRAG is the strongest among them on six of seven base models; HippoRAG~2, ReFoRCE, and CHESS trail in model-dependent order. Every external system stays below 2.0 overall on every base model, while the full \systemname pipeline exceeds 2.5 on five of seven. Failures localize on Tier~R and Tier~M: external systems lack both a routing layer and composable multi-table SQL grounded in concept-to-schema retrieval. Tier~D scores partially recover for the retrieval-based baselines because data-availability disclosure is reachable from text alone; on the Llama models, LightRAG and HippoRAG~2 also recover on Tier~K via text-dominant causal vocabulary.

\paragraph{Routing is universally load-bearing; planning and EKG effects vary by base model.}
Disabling cluster routing drops Tier~M by 0.81--2.67 across all seven base models, the largest single-component drop on Tier~M for every model and consistent with routing's role in anchoring concept-to-schema retrieval. Removing the planning step has highly variable effects: Tier~K drops 3.19 on Qwen~3.6 but only 0.20--0.48 on the other six. Replacing the EKG with Text-RAG drops Tier~M by 1.16--1.28 on Gemini and Qwen~3.6 and 0.50 on DeepSeek, but only 0.04--0.38 on Llama~3.1~8B, Qwen3~8B, and Qwen3~32B (and slightly improves Llama~3.3~70B), suggesting the EKG's typed causal structure delivers most of its benefit when the base model is capable enough to exploit it.

\paragraph{Cross-model heterogeneity in orchestration sensitivity.}
Per-model sensitivity to ablation differs sharply (Figure~\ref{fig:cross_model_ablation}). Qwen~3.6 Flash is most sensitive: overall drops of 2.75 (No Routing) and 2.12 (No Plan) are the two largest single-component drops in the study. DeepSeek~V3.2 is the most tolerant (drops 0.17--0.80 overall); its ReAct baseline alone reaches $\sim$88\% of full-pipeline score, suggesting its intrinsic ReAct loop substitutes for much of what the explicit orchestration adds elsewhere. The four open-source models (Llama~3.1~8B, Qwen3~8B, Qwen3~32B, Llama~3.3~70B) show smaller-magnitude drops (0.10--0.94) that reflect their lower full-pipeline ceilings. The same pipeline is load-bearing on Qwen~3.6, supportive on Gemini, and partially redundant on DeepSeek and the smaller open-source models; any claim that ``pipeline structure matters'' must be qualified by which base model is in scope.

\paragraph{Statistical significance.}
Across all 56 (model, alternative) cells against the full pipeline (7 models $\times$ 8 alternatives: 4 internal ablations + 4 external baselines), \textbf{50 of 56} overall gaps exceed their $\sim$95\% confidence interval computed from the three-seed standard deviations (smallest significant gap: DeepSeek No Plan $+0.17 \pm 0.10$; 46 of the 56 gaps exceed 0.40). The 6 non-significant overall gaps fall on the smaller open-source models; in 5 of them the alternative slightly outperforms Full ($|\Delta| < 0.05$, within seed noise). We use a normal-approximation interval; a paired permutation test over the 75 cases would be tighter. Extending the test to the 224 per-tier Full-vs-alternative gaps (7~$\times$~8~$\times$~4), 168/224 are significant at $p<0.05$ (154/224 under Bonferroni adjustment). The 56 non-significant gaps concentrate on Tier~D (23 of 56) where retrieval-based baselines recover via text disclosure, and on Tier~K (18 of 56) for base models where the planning component delivers less benefit.

\section{Related Work}
\label{sec:related}

\paragraph{Text-to-SQL.}
Text-to-SQL systems on Spider \citep{yu2018spider} and BIRD~\citep{li2023bird}, including LLM-based decomposed-prompting approaches~\citep{pourreza2023dinsql, dong2023c3, gao2024dailsql} and multi-agent frameworks such as MAC-SQL~\citep{wang2024macsql}, assume a small or pre-selected schema. Schema linking~\citep{lei2020reexamine, li2023resdsql}, including recent graph-based work that builds a schema graph from foreign-key and column-name structure to support pathfinding-based linking~\citep{schemaGraphSQL2025, linkAlign2025}, performs string-level, learned-embedding, or schema-graph matching but does not encode domain causal structure. Our concept-to-schema bridge differs in two respects: the graph encodes causal structure (where prior schema-linking work encodes schema-extracted structure), and it is queried for table selection (where prior work targets column-level disambiguation). We frame schema selection as concept-level traversal over a curated knowledge graph that complements downstream text-to-SQL generation.

\paragraph{Graph-augmented retrieval.}
GraphRAG~\citep{edge2024graphrag} and related work~\citep{peng2024graphretrieval, pan2024unifying, zhangSurveyGraphRetrievalAugmented2025, prockoGraphRetrievalAugmentedGeneration2024, zhuKnowledgeGraphGuidedRetrieval2025, lindersKnowledgeGraphextendedRetrieval2025, yangKnowledgeGraphLarge2024} construct knowledge graphs from document corpora and use graph structure to improve passage retrieval for question answering. The \systemname EKG differs from text-derived KGs in three respects: it is expert-curated; it encodes typed causal relations in place of entity co-occurrence; and it mediates retrieval over structured database schemas in place of document passages. The concept-graph layer instantiates the neuro-symbolic pattern advocated by recent reviews~\citep{garcez2023neurosymbolic, kautz2022third}, with the curated graph supplying typed structure and the LLM supplying language understanding and SQL generation.

\paragraph{Disaster AI and operational systems.}
Crisis informatics has produced tools for social media classification~\citep{alam2021crisisbench, imran2015processing}, multimodal damage assessment~\citep{fan2020disaster_city, xiaoCrisiSenseRAGCrisisSensing2026}, and rule-based loss estimation~\citep{fema2019hazus}. Recent surveys of LLMs in disaster management~\citep{leiHarnessingLargeLanguage} note that structured-data querying is underrepresented relative to unstructured text and image analysis, and existing inter-organizational coordination work~\citep{comfort2007interorganizational, bharosa2010challenges} identifies information fragmentation as a primary obstacle. Recent disaster-specific benchmarks~\citep{chenDisastQAComprehensiveBenchmark2026, liuFloodSQLBenchRetrievalAugmentedBenchmark2026, yinDisastIRComprehensiveInformation2025} focus on unstructured-text QA and retrieval over disaster corpora, with concurrent retriever and RAG-system work in the same regime: DMRetriever~\citep{yin2025dmretrieverfamilymodelsimproved} trains a family of dense retrievers tailored to disaster-management text, DisastRAG~\citep{li2026disastragmultisourcedisasterinformation} integrates multi-source disaster information via retrieval-augmented generation, and training-free retrieval-augmented reasoning has been applied to flood-damage nowcasting~\citep{huang2026training}. \systemname is complementary, targeting structured-table QA mediated by a domain-curated concept graph rather than retrieval over disaster text. To our knowledge, no prior computational system routes queries through clusters motivated by ICS functional structure~\citep{fema2017nims, bigley2001ics}.

\section{Conclusion}
\label{sec:conclusion}

We present \systemname, which couples an expert-curated concept-to-schema knowledge graph with a four-stage orchestrator over a 36-table geospatial database. On a 75-case test split, the full system scores in the 1.65--3.56 band across seven base models (Table~\ref{tab:ablation}) and beats four SOTA external baselines by 1.4--2.75$\times$ on every base model (Table~\ref{tab:baselines}), indicating that the concept-to-schema layer plus orchestration matters more than the choice of retrieval substrate or text-to-SQL agent.

The concept-to-schema bridge pattern may transfer to other domains where expert causal knowledge sits alongside time-sensitive structured data, such as medical triage, supply-chain disruption, or infrastructure incident response; a transfer study to a second domain is the natural next step.

\section*{Limitations}

\paragraph{Multimodal extension.}
\systemname currently operates over structured tabular data and the curated EKG. Incorporating complementary multimodal evidence streams such as satellite imagery, sensor telemetry, and crisis-related social media~\citep{xiaoCrisiSenseRAGCrisisSensing2026, alam2021crisisbench} would enrich situational awareness during active incidents and is a promising direction for follow-up work.

\paragraph{Multilingual support.}
The current implementation operates in English. Extending to additional languages, particularly those relevant to multilingual emergency-response contexts in Texas and beyond, would broaden the accessibility of the system and is left to future work.

\section*{Ethics Statement}

\systemname is a decision-support tool for disaster analytics. It is not authoritative incident-command guidance. Its outputs (including any high-criticality recommendations triggered at criticality~$\geq$~3) are intended for review by a qualified emergency-management analyst before any operational action.

\paragraph{Data provenance and licenses.}
All input data sources are public-domain U.S. federal datasets and are used in accordance with their respective terms. FEMA National Risk Index~\citep{fema2023nri} and FEMA Hazus loss-estimation parameters~\citep{fema2019hazus} are released by FEMA as open data with no use restrictions for research. The Homeland Infrastructure Foundation-Level Data (HIFLD) catalog is U.S. government open data, distributed under the federal open-data license. The H3 indexing library is Apache-2.0 (Uber). No personally identifiable information appears in the relational backend; the H3 hex grid aggregates exposure and vulnerability indicators to $\sim$0.74\,km$^2$ cells. \systemname's code, the curated EKG, and the 75-case test split will be released under an Apache-2.0 license upon acceptance.

\paragraph{Model artifacts and terms of use.}
Models accessed via APIs comply with their respective service terms: Gemini~3.1 Flash-Lite Preview and Gemini~2.5 Flash (Google API Terms of Service); DeepSeek V3.2 (open weights under DeepSeek License, with API access via OpenRouter); Qwen~3.6 Flash (Alibaba DashScope service terms); Llama~3.1~8B and Llama~3.3~70B-Instruct (Meta Llama Community License, gated access); Qwen3~8B and Qwen3~32B (Apache 2.0). All models are used for research-only evaluation in this paper. Each model's underlying license permits the academic-comparison use we apply here; we do not redistribute model weights.

\paragraph{Judge-model bias.}
Claim extraction and the Tier~K reasoning judge use a different LLM (Gemini~2.5~Flash) than any pipeline under evaluation, mitigating self-evaluation bias~\citep{zheng2023judging}.

\paragraph{Intended use and misuse risks.}
Intended use: schema-aware question answering over public disaster datasets in research and analyst-assist settings. Out-of-scope uses include autonomous operational decision-making, evacuation-order generation, and any application where the absence of a domain expert in the loop could produce life-safety harm. The dual-use risk most salient to this work is over-reliance: a system that returns confident, well-formatted answers risks displacing expert judgment. The criticality-gated recommendation flag and the data-availability disclosure tier are designed to surface uncertainty explicitly; they are not a substitute for review.

\paragraph{Risks specific to emergency-response deployment.}
Three risks bear naming explicitly given the disaster-analytics application setting. (i)~\textit{Misuse in autonomous decision pipelines.} The system produces structured recommendations gated by a 1--5 criticality score; if these outputs are piped into automated dispatch, resource allocation, or evacuation triggers without human review, errors at higher criticality tiers could cause direct life-safety harm. The criticality-gated tier and the data-availability disclosure surface are designed to flag uncertainty for a human reviewer, not to authorize autonomous action. (ii)~\textit{Automation bias and authority gradient.} Confident, well-formatted analyst output combined with an Incident-Command-styled cluster taxonomy can create a perceived authority that exceeds the system's actual reliability. In time-pressured emergency-management settings, this risk is amplified, and operators may defer to model output rather than apply domain expertise. Operator training on system limits is required before any analyst-assist deployment.

\section*{Acknowledgements}
This work used Grace at Texas A\&M University's High Performance Research Computing (HPRC), and Delta at the National Center for Supercomputing Applications through allocation~\textsc{CIV260030} from the Advanced Cyberinfrastructure Coordination Ecosystem: Services \& Support (ACCESS) program, which is supported by U.S. National Science Foundation grants \#2138259, \#2138286, \#2138307, \#2137603, and \#2138296.

\bibliography{references}

\appendix

\section{Reproducibility Details}
\label{app:reproducibility}

\paragraph{Models.}
The full \systemname pipeline is run with seven base models spanning closed-source and open-source families: Gemini~3.1 Flash-Lite Preview (\texttt{google/\allowbreak gemini-3.1-\allowbreak flash-\allowbreak lite-\allowbreak preview}), DeepSeek~V3.2 (\texttt{deepseek/\allowbreak deepseek-\allowbreak v3.2}), Llama~3.1~8B (\texttt{meta-llama/\allowbreak llama-3.1-8b-\allowbreak instruct}), Qwen3~8B (\texttt{qwen/\allowbreak qwen3-8b}), Qwen3~32B (\texttt{qwen/\allowbreak qwen3-32b}), and Llama~3.3~70B-Instruct (\texttt{meta-llama/\allowbreak llama-3.3-70b-\allowbreak instruct}) all served via OpenRouter, and Qwen~3.6 Flash (\texttt{qwen3.6-flash}) served directly via DashScope. Each (model, condition) cell is run three times with random seeds; Tables~\ref{tab:baselines} and~\ref{tab:ablation} report mean $\pm$ std. Claim extraction and the Tier~K reasoning judge use Gemini~2.5~Flash (\texttt{google/gemini-2.5-flash}), a different model from any pipeline under evaluation, to prevent self-evaluation bias.

\paragraph{Decoding and per-stage hyperparameters.}
All generation calls use \texttt{temperature=0} for determinism. No top-$p$ truncation; default \texttt{max\_tokens}. Per-stage settings: Stage~1 (extract) and Stage~2 (classify) prompts request strict JSON output; for small Llama models (3.1~8B, 3.2~3B, 3.2~1B) we additionally bind \texttt{response\_format=\{"type":"json\_object"\}} to suppress safety-trained refusal of hazard-scenario JSON extraction. Stage~3 (plan) ReAct agent has a tool budget of one \texttt{web\_search} call and one \texttt{query\_knowledge\_graph} call before the plan must be emitted. Stage~4 (execute) ReAct agent has at most 8 \texttt{query\_database} calls and 2 KG lookups; the text-to-SQL sub-pipeline uses a 3-attempt retry loop on validation or execution errors. Per-case execute timeout is 240s (default) or 600s for slower models (Qwen3~32B). LLM-call timeout is 120s. The Tier~K reasoning judge runs with the same \texttt{temperature=0} setting on Gemini~2.5~Flash.

\paragraph{Software stack.}
LangGraph 0.2 for the four-stage orchestrator; LangChain 1.2.15 for the LLM client; Neo4j 5.x for the unified graph (one Docker container, populated from the curated EKG JSON file and the auto-introspected DDCG); DuckDB 1.x for the relational backend (read-only on every query); Tavily Search API for the web-retrieval tool; Python 3.10 in a project-pinned conda environment. The exact environment file, EKG JSON, and orchestration source are released upon acceptance.

\paragraph{Hardware and runtime.}
The pipeline is API-bound: no local GPU is used for inference. Benchmark runs were executed on a single workstation with 32 GB RAM. Wall-clock for a full 75-case run is approximately 60--120 minutes per condition with \texttt{--parallel 3} (three concurrent ReAct workers); the No Routing condition runs longer because every query falls through to a single template that triggers more SQL retries.

\paragraph{Determinism caveat.}
Even at \texttt{temperature=0}, provider routing and load balancing introduce small non-determinism in token-level outputs across runs. To bound this, every (model, condition) cell in Table~\ref{tab:ablation} is run three times with independent random seeds, and we report mean $\pm$ std across the three seeds. The observed std (typically 0.05--0.15 on overall scores, larger on a few cells where seed-level variance is genuinely high) bounds the residual non-determinism.

\paragraph{Data preprocessing.}
The 36 DuckDB tables are built from public-domain federal data sources: per-hex hazard scores from FEMA NRI (riverine flood, hurricane, tornado, wildfire), exposure and population totals from US Census + ACS, social-vulnerability indices from CDC/ATSDR SVI, community resilience from FEMA NRI CRI, and facility inventories from HIFLD (hospitals, fire stations, shelters, power plants). All sources were resampled to the H3 resolution-8 hexagonal grid (a one-time preprocessing pass, $\sim$2 hours wall-clock on a workstation) and joined on the shared \texttt{hex\_id} key. The DDCG node graph is auto-introspected from the resulting DuckDB at system startup; no manual schema curation is involved. The exact preprocessing scripts, intermediate tables, and DuckDB build commands are released upon acceptance.

\paragraph{Compute budget.}
The full evaluation campaign spans approximately 225 benchmark cells: 7 base models $\times$ 5 ablation conditions $\times$ 3 seeds (105 cells), 7 base models $\times$ 4 external baselines $\times$ 3 seeds (84 cells, one permanently dropped due to context-length overflow), and 4 further ablation conditions explored on 3 base models during system design (36 cells). Aggregate token usage is on the order of 700--900M input tokens and 80--130M output tokens on the pipeline models, plus $\sim$70M tokens on the Gemini~2.5~Flash claim extractor across all evaluation runs. Total billed cost was approximately USD~500--700 at the OpenRouter, DashScope, and Google API rate sheets at the time of the experiments. No GPU compute was used.

\section{Data Statistics}
\label{app:data_stats}

\paragraph{Test-split composition.}
The frozen test split contains 75 cases organized across the four tiers from Table~\ref{tab:tiers}: R (routing, $n$=17), K (EKG grounding, $n$=19), M (multi-table composition, $n$=26), and D (data-availability disclosure, $n$=13). Scope splits 60 county-scoped queries against 15 statewide queries; the statewide subset spans all four tiers and is on average $\sim$0.5 judge-score points harder per tier than the county-scoped subset. Hazard categories are distributed across tiers: flood/hurricane queries account for the largest share ($\sim$45\%) reflecting the dominant Texas hazard profile, followed by wildfire ($\sim$18\%), tornado ($\sim$15\%), multi-hazard compound scenarios ($\sim$12\%), and power-disruption / drought / other ($\sim$10\%). All cases reference Texas geographies; per-case county and hazard tags are released alongside the benchmark file.

\paragraph{Construction protocol.}
The test split was authored in two phases by the paper authors. Phase~1 created 60 county-scoped cases against the catalog of available tables and the curated EKG concepts, with a tier-balanced design (R/K/M/D quotas met before any case-level review). Phase~2 added 15 statewide multi-county cases to introduce compound-aggregation difficulty (e.g., Texas Panhandle, Permian Basin, Gulf Coast). All cases were reviewed against the DDCG schema to confirm answerability under the available tables. The split was frozen before any pipeline execution and was not opened during system development; a separate, equally sized development split was used for prompt iteration and ablation prototyping, and its scores are not reported.

\paragraph{Gold-fact extraction and annotator details.}
Tier R gold labels (correct ICS cluster and query-type identifier) and Tier D gold labels (which data warnings should fire) are deterministic and derived from the test-case authorship. Tier K causal-token gold labels are extracted directly from the curated EKG along causal-path traversals for each case; this design measures EKG retrieval grounding rather than independent causal correctness. Tier M numeric gold values are produced by running canonical reference SQL against the live DuckDB and capturing the result; tolerance bands ($\pm$100 hex counts, $\pm$10\% on averaged quantities) absorb minor numeric variance from equivalent SQL formulations. The 20-case sample (5 per tier) used to validate claim-extraction precision was annotated by one author (single-rater); the per-claim-type precision rates are 97\% numeric, 97\% causal, 85\% boolean, with overall 95\% claim-extraction precision. The 20-case sample IDs and annotation notes are released alongside the benchmark.

\paragraph{Geographic coverage and known biases.}
All 75 cases reference Texas geographies. The split is intentionally Texas-only because the underlying DuckDB build is Texas-only at H3 resolution~8. As a result, the benchmark does not measure cross-region generalization, hazard portfolios specific to other U.S. regions (e.g., Pacific Northwest seismic, Atlantic Coast nor'easter), or international disaster contexts. Cross-region extension is identified as a primary future-work direction in Limitations.

\section{Internal-Ablation Cross-Model Figure}
\label{app:cross_model_figures}

The bar chart below visualises overall scores from Table~\ref{tab:ablation} across all seven base models. It is referenced from \S\ref{sec:results}; the corresponding external-baseline view is shown in Figure~\ref{fig:cross_model_baselines}.

\begin{figure*}[!htbp]
  \centering
  \includegraphics[width=0.5\textwidth]{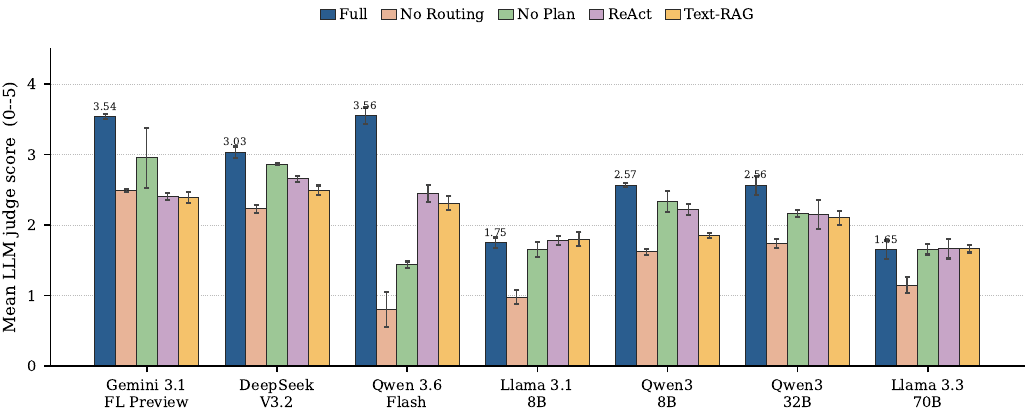}
  \caption{Overall mean LLM judge score for the full \systemname pipeline versus four internal-component ablations (No Routing, No Plan, ReAct, Text-RAG) across all seven base models. DisasterLex Full bars are value-labeled. Bars are means over three random seeds; error bars are $\pm$ one standard deviation.}
  \label{fig:cross_model_ablation}
\end{figure*}

\section{Pipeline Prompt Structure}
\label{app:prompts}

The four orchestration stages (\S\ref{sec:pipeline}) use structured prompt templates; condensed verbatim excerpts of Stages~1, 2, and~4 are shown in Table~\ref{tab:prompts}. Stage~3 (causal-informed planning) is a ReAct~\citep{yao2023react} agent with two tools, \texttt{query\_knowledge\_graph} (retrieves causal edges from the EKG by concept ID) and \texttt{web\_search} (Tavily). The agent decomposes the query into 2--4 sub-questions, calls the KG tool to retrieve causal context (at most 2 calls), optionally calls \texttt{web\_search} for current events, and emits a structured plan with named SQL targets and expected metrics; the plan is consumed as text by Stage~4. The full prompt text for all stages (including all rule clauses) is released alongside the code.

\begin{table*}[!t]
  \centering
  \small
  \renewcommand{\arraystretch}{1.15}
  \begin{tabular}{p{0.96\textwidth}}
    \toprule
    \textbf{Stage 1: Context Extraction} \\
    \midrule
    ``Analyze the disaster management query below. Extract \texttt{area\_of\_interest}, \texttt{hazard\_type} (one of \texttt{flood}, \texttt{hurricane}, \texttt{wildfire}, \texttt{tornado}, \texttt{drought}, \texttt{power\_disruption}, \texttt{multi\_hazard}), \texttt{criticality} (1--5), and a short rationale.\par
    Criticality: 1=Monitor (routine planning); 2=Advisory (watch); 3=Action (warning/imminent, `in N hours'); 4=Urgent (active impact: `is currently affecting', `has struck'); 5=Critical (`Cat 5', `catastrophic', `mass casualty'). Any `Category 5' / `Cat 5' hurricane always assigns 5. If the query is past-tense / what-if / planning, cap criticality at 3.\par
    User Query: `\{query\}'. Return ONLY JSON: \texttt{\{"area\_of\_interest":}~\ldots, \texttt{"hazard\_type":}~\ldots, \texttt{"criticality":}~N, \texttt{"criticality\_rationale":}~\ldots\texttt{\}}'' \\
    \midrule
    \textbf{Stage 2: Cluster Routing} \\
    \midrule
    ``Determine which operational cluster and query type best fit this request.\par
    User Query: `\{query\}'. Context: \{hazard\} in \{area\} (Criticality \{c\}).\par
    Cluster Descriptions: (1)~\texttt{life\_safety\_operations}: OPERATIONAL queries: active rescue, SAR-zone identification, evacuation, medical triage. (2)~\texttt{damage\_assessment\_response}: ANALYTICAL queries: post-event damage counts, economic loss, recovery planning, resource gaps, AND causal mechanism questions. (3)~\texttt{infrastructure\_mitigation}: protecting infrastructure before/during an event: hardening, surge exposure to critical facilities, power-grid vulnerability.\par
    Available Query Types per Cluster: \{enumerated list\}. Disambiguation: `identify facilities in <risk zone>' $\rightarrow$ infrastructure\_mitigation; `identify populations with compound risk (no deployment)' $\rightarrow$ damage\_assessment\_response; `identify SAR / shelter / evacuation zones' $\rightarrow$ life\_safety\_operations.\par
    Return ONLY JSON: \texttt{\{"cluster":\ldots, "query\_type":\ldots\}}'' \\
    \midrule
    \textbf{Stage 4: Tool-Augmented Execution} \\
    \midrule
    ``You are an Incident Command analyst. Execute the provided analysis plan using actual data.\par
    CRITICALITY: \{c\}. AREA: \{area\}. \{data-availability disclosure block, if any\}. \{routing template body for the chosen \texttt{(cluster, query\_type)}\}. \{concept-aware schema hints\}.\par
    Constraints: at most 8 \texttt{query\_database} calls (use CTEs / JOINs to combine); never call schema-introspection (\texttt{DESCRIBE}, \texttt{PRAGMA}, \texttt{information\_schema}, etc.), query actual data tables directly; at most 2 KG lookups; if SQL retries exceed 3 attempts on one logical query, abandon and answer with available data.\par
    Recommendations gated by criticality: 1--2 informational; 3 active-response; 4--5 life-safety missions.\par
    Self-check before finalizing: (a)~did I state every primary numeric metric the question asked for, with its exact value? (b)~If the question mentions equity / vulnerable populations, did I address that dimension? (c)~If the question asks `how/why/cascade/driver', did I name the causal chain using KG concept IDs (e.g.\ \texttt{flood\_occurrence INCREASES power\_disruption REDUCES hospital\_operations})? (d)~Are my recommendations gated by Criticality? (e)~For regional aggregates, did I sum across the FULL named scope?\par
    Output ends with a MANDATORY structured-facts block (one snake-case line per aggregated metric the question requested).'' \\
    \bottomrule
  \end{tabular}
  \caption{Prompt templates for the four-stage orchestration pipeline (\S\ref{sec:pipeline}). Stage~3 (causal-informed planning) is described in prose above; its ReAct system prompt is released alongside the code. Curly braces denote runtime-substituted placeholders (e.g., \texttt{\{query\}}, \texttt{\{hazard\}}, \texttt{\{area\}}, \texttt{\{c\}}).}
  \label{tab:prompts}
\end{table*}

\section{EKG Schema Details}
\label{app:ekg}

The curated Expert Knowledge Graph contains 107 concept nodes distributed across eight categories, connected by 117 typed causal edges (Table~\ref{tab:ekg_schema}). The three concept-to-schema bridge target categories are derived from the auto-introspected DDCG (36 \texttt{DataTable} nodes, 150 \texttt{DataColumn} nodes, 7 \texttt{JoinRule} nodes).

\begin{table}[!htbp]
  \centering
  \small
  \begin{tabular}{l c}
    \toprule
    \textbf{Node category} & \textbf{Count} \\
    \midrule
    Feature          & 19 \\
    Intermediate     & 56 \\
    Outcome          & 4 \\
    Domain anchor    & 16 \\
    Intervention     & 5 \\
    Exposure         & 3 \\
    Response asset   & 2 \\
    Infrastructure   & 2 \\
    \midrule
    \textbf{Total concepts} & \textbf{107} \\
    \bottomrule
  \end{tabular}
  \quad
  \begin{tabular}{l c}
    \toprule
    \textbf{Edge type} & \textbf{Count} \\
    \midrule
    \textsc{Increases} & 56 \\
    \textsc{Reduces}   & 49 \\
    \textsc{Indicates} & 9 \\
    \textsc{Requires}  & 2 \\
    \textsc{Scales}    & 1 \\
    \midrule
    \textbf{Total causal} & \textbf{117} \\
    \textbf{\textsc{Maps\_To}} & \textbf{52} \\
    \bottomrule
  \end{tabular}
  \caption{EKG schema breakdown. Concept categories follow the \texttt{type} field on each node in the curated EKG; edge types are stored on each \texttt{Concept}--\texttt{Concept} relationship. The 52 concept-to-schema edges connect concept nodes to \texttt{DataTable} nodes that are loaded in the current DuckDB build.}
  \label{tab:ekg_schema}
\end{table}

\paragraph{Sample concept-to-schema edges.}
\texttt{flood\_\allowbreak occurrence} maps to \texttt{HP\_FLD\_002} (NRI riverine flood risk); \texttt{vulnerability} maps to \texttt{VUL\_002} (Social Vulnerability Index) and \texttt{VUL\_004} (population social vulnerability index); \texttt{shelters} maps to \texttt{HIFLD-EMERGENC-SHELTER-N} (HIFLD national shelter system facilities); \texttt{hospitals} maps to \texttt{EX\_LIFE\_004} (hospital facility counts); \texttt{population} maps to \texttt{EX\_POP\_001} (population per hex). The remaining 26 bridges follow the same one-concept-to-one-or-two-tables pattern.

\section{ICS Mapping}
\label{app:ics_mapping}

\begin{table}[!htbp]
  \centering
  \small
  \renewcommand{\arraystretch}{1.2}
  \setlength{\tabcolsep}{4pt}
  \begin{tabular}{@{}>{\raggedright\arraybackslash}p{0.40\columnwidth} >{\raggedright\arraybackslash}p{0.54\columnwidth}@{}}
    \toprule
    \textbf{\systemname element} & \textbf{ICS analogue} \\
    \midrule
    \multicolumn{2}{@{}l}{\textbf{Routing clusters}} \\
    Life-safety operations       & Operations Section (Medical \& Rescue) \\
    Damage assessment \& response & Planning Section (Situation Unit) feeding Operations \\
    Infrastructure mitigation    & Planning (Resources Unit) + Logistics (Service Branch) \\
    \addlinespace
    \multicolumn{2}{@{}l}{\textbf{Cross-cutting gates}} \\
    Criticality gate ($\geq 3$)  & Incident Action Plan urgency tier (life-safety priority) \\
    Data-availability disclosure & Documentation Unit (Planning Section) \\
    \bottomrule
  \end{tabular}
  \caption{Mapping from \systemname pipeline elements to ICS functional sections.}
  \label{tab:ics_mapping}
\end{table}
The three routing clusters in \S\ref{sec:pipeline} are loosely motivated by the functional structure of the Incident Command System~\citep{fema2017nims, bigley2001ics}; they are not formally aligned with it. Table~\ref{tab:ics_mapping} gives the explicit correspondence. The mapping compresses the five ICS functional sections (Command, Operations, Planning, Logistics, Finance/Administration) into the subset relevant for analyst-side decision support and omits Command and Finance/Administration as out of scope; the routing decision is therefore best read as a coarse functional triage at the cluster level.

\section{Evaluation Conditions}
\label{app:conditions}

\begin{table}[!t]
  \centering
  \small
  \renewcommand{\arraystretch}{1.2}
  \begin{tabularx}{\columnwidth}{@{}l >{\raggedright\arraybackslash}X@{}}
    \toprule
    \textbf{Condition} & \textbf{Mechanism} \\
    \midrule
    Full        & Concept retrieval + ICS routing + planning + execution. \\
    No Routing  & Hardcoded cluster template; area extracted via regex. \\
    No Plan     & No pre-execution causal research step. \\
    ReAct       & ReAct agent~\citep{yao2023react}; same tools, no routing/template/plan. \\
    Text-RAG    & Source corpus as text chunks over a ReAct backbone. \\
    LightRAG    & Auto-extracted entity-relation graph (hybrid retrieval) over a ReAct backbone~\citep{guoLightRAGSimpleFast2025}. \\
    \bottomrule
  \end{tabularx}
  \caption{The five baseline and ablation conditions evaluated against the full pipeline (referenced from \S\ref{sec:conditions}).}
  \label{tab:conditions}
\end{table}

\section{External Baseline Configuration}
\label{app:baselines}

\paragraph{LightRAG.}
We index the same source corpus that backs the EKG (TDIS chunks; $n$=2{,}577) using a recent release of the upstream LightRAG implementation~\citep{guoLightRAGSimpleFast2025}, producing an on-disk index of 96~MB containing 85 entities, 85 relations, and 926 chunks. LLM extraction calls use Gemini~3.1 Flash-Lite Preview via OpenRouter; embeddings use \texttt{sentence-\allowbreak transformers/\allowbreak all-MiniLM-L6-v2} (384-dim) to avoid an OpenAI dependency for retrieval. At query time, we retrieve in \texttt{hybrid} mode (local entity retrieval + global community retrieval) with \texttt{top\_k=10}, surfacing the resulting context block as the sole tool response inside an otherwise unmodified ReAct backbone. The exact commit hash and index artifacts are released upon acceptance.

\paragraph{Author's configuration.}
LightRAG is configured under its authors' recommended defaults except for the unavoidable embedding-model swap (\texttt{sentence-\allowbreak transformers/\allowbreak all-MiniLM-L6-v2} in place of an OpenAI embedding model). We did not fine-tune LightRAG's KG extraction on a disaster-specific ontology and did not provide it with the EKG. Such adaptations would amount to porting the EKG into LightRAG's architecture, which is the contribution this paper is testing (see~\S\ref{sec:conclusion}). LightRAG is evaluated on all seven base models with three seeds each.

\section{Worked Example}
\label{app:example}

We trace the full pipeline through a representative Tier~M case (\texttt{draft\_m33}) to illustrate concept-aware schema retrieval and the per-case scoring rubric in action. All field values shown below are taken verbatim from the released benchmark file and the Full-pipeline result for Gemini~3.1 Flash-Lite Preview (seed~1).

\paragraph{Input.}
``Flooding in Hidalgo County. Triage the medical system under flood conditions: which hospitals are at flood risk, and identify SAR priority zones where patients may be stranded.''

\paragraph{Stages~1--2 (extract + route).}
Table~\ref{tab:worked_routing} shows the routing state populated by the extractor and the cluster/template selected by the router (1 of 18 prompt templates).

\begin{table}[!h]
  \centering
  \small
  \begin{tabular}{@{}l l@{}}
    \toprule
    \textbf{Field} & \textbf{Value} \\
    \midrule
    \texttt{area\_of\_interest} & Hidalgo County \\
    \texttt{hazard\_type}       & flood \\
    \texttt{criticality}        & 4 (active impact event) \\
    \texttt{cluster}            & \texttt{life\_safety\_operations} \\
    \texttt{query\_type}        & \texttt{medical\_triage} \\
    \bottomrule
  \end{tabular}
  \caption{Pipeline state after Stages~1--2 for \texttt{draft\_m33}. The selected prompt template's stated objective is to ``assess the vulnerability of the healthcare network and identify potential medical deserts during an active event.''}
  \label{tab:worked_routing}
\end{table}

\paragraph{Stage~3 (plan).}
Causal context retrieved from the EKG describes the cascade flood occurrence \textsc{Increases} road access disruption $\rightarrow$ staff and supply chain isolation $\rightarrow$ backup power exhaustion $\rightarrow$ hospital operations failure. Plan: enumerate Hidalgo County hex cells, count hospital-bearing hexes, intersect with high-flood-risk hexes (riverine flood score $\geq$~75), and identify SAR priority zones.

\paragraph{Stage~4 (execute).}
Concept-aware schema retrieval activates the flood-occurrence, hospitals, and community-resilience concepts. Concept-to-schema traversal selects four tables: \texttt{HP\_FLD\_002} (flood score), \texttt{EX\_LIFE\_004} (hospital count), \texttt{CR\_001} (community resilience), and the county crosswalk. The injected schema context is 14 columns; full-schema injection would expose 150 columns. The generated SQL joins on the hex identifier, filters on county equal to ``Hidalgo County'', applies the high-flood-risk threshold, and returns the affected hex sets together with five operational recommendations (mutual aid activation, field-hospital staging, FEMA Preliminary Damage Assessment, generator fuel resupply, SAR prioritization). The synthesized answer reports \textbf{12 total hospital hexes}, \textbf{6 high-risk hospital hexes (50\%)}, and includes the causal chain from Stage~3.

\paragraph{Scoring.}
The benchmark file specifies five deterministic checks for \texttt{draft\_m33} (Tier~M total weight 5.0; no LLM reasoning judge). Table~\ref{tab:worked_scoring} shows each check, its expected and actual value, and the weighted contribution to the per-case score.

\begin{table}[!h]
  \centering
  \footnotesize
  \setlength{\tabcolsep}{3pt}
  \begin{tabular}{@{}l l c l c@{}}
    \toprule
    \textbf{Check} & \textbf{Kind} & \textbf{W} & \textbf{Expected\ / Actual} & \textbf{+} \\
    \midrule
    routing      & pipeline match     & 1.0 & match              & 1.00 \\
    num1         & numeric ($\pm 3$)  & 1.5 & 12.0 / 12.0        & 1.50 \\
    num2         & numeric ($\pm 50$) & 1.5 & 2597.0 / missing   & 0.00 \\
    bool         & boolean            & 0.5 & true / true        & 0.50 \\
    rec.\ count  & count $\ge 2$      & 0.5 & 2 / 5              & 0.50 \\
    \midrule
    \multicolumn{4}{r}{\textbf{Fact score}}                         & \textbf{3.50} \\
    \multicolumn{4}{r}{Reasoning score (Tier~M: none)}              & 0.00 \\
    \multicolumn{4}{r}{\textbf{\texttt{judge\_score\_raw}}}         & \textbf{3.50} \\
    \bottomrule
  \end{tabular}
  \caption{Deterministic scoring breakdown for case m33. W is the check weight; + is the weighted contribution (W $\times$ score fraction). Tier~M total is 5.0. The pipeline recovers four of five checks; the missing SAR-priority hex count is the single point of failure. This per-case score contributes one cell to the Tier~M column of Tables~\ref{tab:baselines} and~\ref{tab:ablation}.}
  \label{tab:worked_scoring}
\end{table}

\section{Tier~K Reasoning-Judge Prompt}
\label{app:tierk_judge}

For the 5 statewide Tier~K cases, a separate LLM judge (Gemini~2.5~Flash) scores the reasoning quality of the system's answer on a 0--1 scale per check, contributing up to 1 point of the per-case 1--5 total. The judge receives a JSON payload containing the question, the deterministic gold facts, the system answer, and a list of case-specific reasoning checks (each with an instruction and a list of \texttt{required\_points} drawn from the curated EKG). The condensed template is shown in Table~\ref{tab:tierk_prompt}.

\begin{table}[!h]
  \centering
  \small
  \renewcommand{\arraystretch}{1.15}
  \begin{tabularx}{\columnwidth}{@{}X@{}}
    \toprule
    \textbf{Tier~K Reasoning-Judge Prompt} \\
    \midrule
    ``Evaluate the system answer against each reasoning check. Return valid JSON only as \texttt{\{"checks": [\{"id": "\ldots", "score\_fraction": 0.0, "reason": "\ldots"\}]\}}. Score fractions must be between 0 and 1.''\par
    \{JSON payload with the following keys\}\par
    \texttt{question}: the user query.\par
    \texttt{gold\_facts}: the deterministic gold facts for the case.\par
    \texttt{system\_answer}: the system's answer text.\par
    \texttt{checks}: a list of reasoning checks. Each check contains:\par
    \quad \texttt{id}: identifier for this check.\par
    \quad \texttt{kind}: check kind (e.g., \texttt{reasoning\_judge}).\par
    \quad \texttt{prompt}: case-specific reasoning instruction (e.g., ``Evaluate whether the answer explains the physical mechanism by which impervious surface coverage increases stormwater runoff and thereby increases flood occurrence. The reasoning should reference county-specific data, not be purely generic.'').\par
    \quad \texttt{required\_points}: list of expected points drawn from the curated EKG (e.g., (1)~explains that impervious surfaces prevent infiltration, increasing surface runoff; (2)~connects increased runoff to higher flood risk using data; (3)~recommendations reference the causal insight). \\
    \bottomrule
  \end{tabularx}
  \caption{Tier~K reasoning-judge prompt template (\S\ref{sec:benchmark}). The judge (Gemini~2.5~Flash) receives a JSON payload with the question, gold facts, system answer, and a list of case-specific reasoning checks, and returns one \texttt{score\_fraction} $\in [0, 1]$ per check together with a short rationale.}
  \label{tab:tierk_prompt}
\end{table}

The reasoning score is added to the deterministic fact-check score (0--4 points: numeric facts, boolean claims, entity-set matches, routing-state verification) to produce the per-case 1--5 judge score reported in Table~\ref{tab:ablation}. For Tier~R, M, and D cases, no LLM-judged reasoning component is used; those cases are scored purely by deterministic checks.

\section{Annotation Process for the 20-Case Claim-Extraction Validation}
\label{app:annotation}

We validated the LLM claim-extraction pipeline (Gemini~2.5~Flash) on a 20-case sample drawn uniformly from the four tiers (5 cases per tier). One author served as the annotator. For each sampled case, the annotator received: (i)~the question, (ii)~the system answer, (iii)~the LLM-extracted claim list, and was asked to mark each extracted claim as \textit{correct} (the claim is supported by the system answer text), \textit{incorrect} (the claim is not supported, including hallucinated numbers or causal edges), or \textit{ambiguous} (the answer text is unclear). Precision is reported as $\text{correct} / (\text{correct} + \text{incorrect})$ over all extracted claims in the sample; ambiguous claims are excluded from the denominator.

The 20 cases yielded 247 extracted claims in total. Per-claim-type precision: numeric ($n$=109, 97\%), causal ($n$=63, 97\%), boolean ($n$=52, 85\%), entity-set ($n$=23, 95\%). The lower boolean precision is driven by the LLM's tendency to convert hedged answer phrases (``most hospitals'', ``a few zones'') into stricter boolean claims, which the annotator marked incorrect when the underlying answer text did not commit to a definite truth value. We discuss the single-rater limitation in Limitations; inter-annotator agreement statistics are not reported because the validation was performed by a single annotator. A second-rater pass on the same 20 cases is planned for a future revision.

\section{Per-Tier Example Cases}
\label{app:tier_examples}

To make the tier taxonomy concrete, Table~\ref{tab:tier_examples} shows one representative case from each of the four tiers (R, K, M, D) drawn from the 75-case test split. Full case IDs let the released benchmark file be cross-referenced.

\begin{table}[!h]
  \centering
  \small
  \renewcommand{\arraystretch}{1.2}
  \begin{tabularx}{\columnwidth}{@{}l >{\raggedright\arraybackslash}X@{}}
    \toprule
    \textbf{Tier/Case ID} & \textbf{Question (truncated)} \\
    \midrule
    R \texttt{draft\_r12} & Active wildfire is burning in Kerr County. Map the highest-risk fire zones, identify transmission infrastructure in the fire path, and estimate how many buildings are in the threat area. \\
    K \texttt{draft\_k13} & Severe flooding in Webb County. How does social vulnerability compound emergency response challenges during an active flood? Identify the highest-priority zones for swift-water rescue deployment. \\
    M \texttt{draft\_m18} & Tornado warning in Midland County. Triage critical lifeline infrastructure: count hospital hexes and substation hexes in high tornado risk zones. Report resource gap. \\
    D \texttt{draft\_d09} & Hurricane impact in Cameron County. Assess social vulnerability across all hexes, including those with missing SoVI data, and report the coverage fraction. \\
    \bottomrule
  \end{tabularx}
  \caption{One example case per tier (lowest-numbered case of each tier in the test split). The Tier~D example deliberately asks for an aggregate that crosses hexes with missing SoVI data; the system is expected to report the coverage fraction.}
  \label{tab:tier_examples}
\end{table}

\section{Failure-Mode Notes for Tier~M}
\label{app:failure_modes}

Tier~M (multi-table composition, $n$=26) is the lowest-scoring tier under the full pipeline on every base model (Table~\ref{tab:ablation}). Qualitatively, the failure modes we observe most often on Tier~M cases (without a labeled quantitative breakdown) are: (i)~\textit{partial schema selection}: concept matching activates the right tables but misses one of the auxiliary tables needed to complete a 3- or 4-table join (e.g., includes shelters and surge exposure but omits the crosswalk needed to filter by county); (ii)~\textit{join-key errors}: the LLM joins on \texttt{hex\_id} but applies a county filter via the wrong table, producing zero rows or incorrect aggregates; (iii)~\textit{aggregation drift}: when a question asks for both a per-hex count and a regional sum, the system reports one or the other but not both; (iv)~\textit{tolerance-band misses}: numeric answers fall outside the $\pm$100 hex-count band on cases where the system uses a slightly different filter than the gold. A formal annotation of all Tier~M failures by category is left to a future revision; the per-case error logs are released alongside the benchmark for independent re-analysis.

\end{document}